\documentclass[preprint,12pt,5p,times,twocolumn]{elsarticle}

\usepackage{amsmath,amssymb,amsfonts}
\usepackage{algorithmic}
\usepackage{graphicx}
\usepackage{algorithm,algorithmic}
\usepackage{hyperref}
\hypersetup{hidelinks=true}
\usepackage{textcomp}

\usepackage{textcomp}
\usepackage{multirow}
\usepackage{comment}
\usepackage[super]{nth}
\usepackage [latin1]{inputenc}
\usepackage[table]{xcolor} 
\usepackage{booktabs} 

\usepackage{pgfplots}
\usetikzlibrary{calc,positioning, decorations.pathreplacing}
\pgfplotsset{compat=1.18}

\usepackage{subcaption}
\usepackage{multicol}
\usepackage{tabularx}
\usepackage{makecell}

\begin{document}

\begin{frontmatter}



\title{VAE-IF: Deep feature extraction with averaging for fully unsupervised artifact detection in routinely acquired ICU time-series}


\author[label1,label5]{Hollan Haule} 
\author[label2]{Ian Piper} 
\author[label4]{Patricia Jones} 
\author[label3]{Chen~Qin} 
\author[label2]{Tsz-Yan Milly Lo} 
\author[label1]{Javier~Escudero} 

\affiliation[label1]{organization={Institute for Imaging, Data and Communications (IDCOM), School of Engineering, University of Edinburgh},
            city={Edinburgh},
            postcode={EH9~3FB}, 
            country={U.K.}}

\affiliation[label2]{organization={Centre of Medical Informatics, Usher Institute, University of Edinburgh},
            city={Edinburgh},
            country={U.K.}}
\affiliation[label3]{organization={Department of Electrical and Electronics Engineering \& I-X, Imperial College London},
            city={London},
            country={U.K.}}
\affiliation[label4]{organization={Department of Child Life and Health, University of Edinburgh},
            city={Edinburgh},
            country={U.K.}}    

\affiliation[label5]{organization={Corresponding author},
            addressline={hhaule@ed.ac.uk}}
            
\begin{abstract}
Artifacts are a common problem in physiological time series collected from intensive care units (ICU) and other settings. They affect the quality and reliability of clinical research and patient care. Manual annotation of artifacts is costly and time-consuming, rendering it impractical. Automated methods are desired. Here, we propose a novel fully unsupervised approach to detect artifacts in clinical-standard, minute-by-minute resolution ICU data without any prior labeling or signal-specific knowledge. Our approach combines a variational autoencoder (VAE) and an isolation forest (IF) into a hybrid model to learn features and identify anomalies in different types of vital signs, such as blood pressure, heart rate, and intracranial pressure. We evaluate our approach on a real-world ICU dataset and compare it with supervised benchmark models based on long short-term memory (LSTM) and XGBoost and statistical methods such as ARIMA. We show that our unsupervised approach achieves comparable sensitivity to fully supervised methods and generalizes well to an external dataset. We also visualize the latent space learned by the VAE and demonstrate its ability to disentangle clean and noisy samples. Our approach offers a promising solution for cleaning ICU data in clinical research and practice without the need for any labels whatsoever.
\end{abstract}












\end{frontmatter}



\section{Introduction}
\label{introduction}
Continuous monitoring of vital signs in Intensive Care Units (ICU) is essential for patient management and treatment. These vitals typically include signals such as blood pressure, heart rate, and temperature. The clinical practice in most ICU involves the routine acquisition of these signals at low temporal resolution (minute-by-minute samples). This data can become a very useful resource for  medical research, as it could advance knowledge about disease processes and patient care \cite{Depreitere2016}. However, extracting useful insights from this data is a big challenge due to artifacts.

Routine collection of ICU data is done in uncontrolled conditions. The focus is clinical care rather than medical research \cite{Shillan2019UseReview}, which often affects  data quality. By way of illustration, probe changes and accidental probe displacement, position changes, and clinical interventions \cite{Johnson2016} can manifest as ``spikes'' or ``flat-line'' artifacts in the collected data, among other forms of artifacts. Therefore, re-purposing this data for offline clinical research risks getting misleading results if artifacts are not dealt with.

The gold standard for dealing with data artifacts is manual annotation by experienced researchers, but this is costly and non-scalable \cite{Alkhachroum2020}. Therefore, developing techniques to help automate this task is an active and important area of research. There are several approaches for detecting artifacts in diverse types of physiological time series. Some of these approaches include Signal Quality Indices (SQI) \cite{Yaghouby2017VariabilityStudy, Silva2011PhotoplethysmographFiltering, Li2008, Sun2006, Zong2004} based on template matching, frequency-based methods \cite{Saeed2000}, and machine learning (ML) \cite{Tsien2001, Mataczynski2022End-to-EndLearning, Subramanian2021UnsupervisedActivity, Chen2021SignalApproaches, Lee2020ArtifactInjury, Wu2020, Edinburgh2019}. Unsupervised ML techniques are of special interest in this domain because of the scarcity of labeled data. However, existing approaches either are signal-specific or work only on high-temporal resolution, quasi-periodic physiological signals, which leaves a gap for methods that work on non-periodic, routinely collected physiologic data from ICU clinical practice.

We aim to develop unsupervised ML methods for automated artifact detection in minute-by-minute ICU time series. Building upon a labeled real-world dataset, we are able to implement supervised ML techniques known for their reliable performance with labeled data as benchmarks. Importantly, we introduce a novel \textit{completely unsupervised} ML approach that exhibits comparable performance to the fully supervised benchmarks. We validate the generalizability of this approach on recordings from an external dataset. Our methods leverage concepts from LSTMs, variational autoencoders (VAE), and decision trees. In particular, our contributions are:
\begin{enumerate}
    \item A completely unsupervised hybrid model to detect artifacts in routinely collected ICU data samples based on VAE and isolation Forest model (VAE-IF) that does not require any previous labeling of artifacts and exploits averaged latent sample representations for enhanced accuracy.
    \item Implementations of supervised benchmarking models (including deep learning and traditional statistical methods) and comparisons of the performance of our VAE-IF model to that of the fully supervised models for artifact detection in ICU data.
    \item Experiments with real world data to explore the learned latent space of VAE-IF, demonstrating its ability to disentangle artifacts from clean data despite being unsupervised.
\end{enumerate}

The manuscript is organised as follows. Section~\ref{litrev} covers literature review. Section~\ref{methodology} introduces our methodology and benchmark techniques. Section~\ref{results} then presents the experiments and results. Finally, Sections~\ref{discussion} to \ref{conclusion} discuss and conclude our study.


\section{Literature review}\label{litrev}
Some of the earliest works on artifact detection and removal in ICU physiological time series data used a combination of statistical rule-based methods \cite{Jakob2000} and wavelet transform \cite{Saeed2000}. Jakob \MakeLowercase{\textit{et al.}} \cite{Jakob2000} compared Rosner statistic, median detection, and a rule-based slope detection method to detect artifacts in physiological data from a paediatric ICU. On average, the three methods achieved sensitivity of 66\%, 72\%, and 64\% respectively. The authors concluded that median filtering is efficient against spikes/outliers but not suitable for artifacts spanning a long duration. Saeed \MakeLowercase{\textit{et al.}} \cite{Saeed2000} used the Haar wavelet transform to detect artifacts and physiological trend changes in multivariate physiological signals. The method used normalised wavelet coefficients to detect significant trend changes in the signals. Using scale 4, which represented changes in 160-second intervals, and XOR logic the method could detect physiological implausible changes as artifacts. 12 out of 15 segments containing artifacts were detected with no false positives. However, the method relies on trend changes and may make it ineffective against flatline artifacts.

SQI is among the most prominent techniques encountered in the literature about physiological time series artifact detection and removal \cite{Yaghouby2017VariabilityStudy, Silva2011PhotoplethysmographFiltering, Li2008, Sun2006, Zong2004}. SQI estimates signal quality based on the morphology of a pulsatile physiological signal. SQI helps identify and exclude segments corrupted by noise when deriving vital signs. Zong \MakeLowercase{\textit{et al.}} \cite{Zong2004a} developed an SQI combining waveform morphological features and fuzzy logic to estimate the quality of arterial blood pressure (ABP) waveform, a continuous value ranging between 0 and 1, which was then used to estimate and reduce blood pressure false alarms. Sun \MakeLowercase{\textit{et al.}} \cite{Sun2006} developed a binary-valued SQI for ABP combining morphological features and physiologic constraints, where 0 and 1 indicate clean and noisy beats, respectively. Apart from SQI for blood pressure signal, Yaghouby \MakeLowercase{\textit{et al.}} \cite{Yaghouby2017VariabilityStudy} developed an SQI for electrocardiogram signals that uses a correlation coefficient between a beat detected in a segment and template beat as an indicator for signal quality. Silva \MakeLowercase{\textit{et al.}} \cite{Silva2011PhotoplethysmographFiltering} developed an SQI for multi-channel physiologic recordings using a multi-channel adaptive filter and applied it to  photoplethysmogram (PPG) signal. Though this SQI offers some flexibility, it is important to note that, in general, SQI are tuned to work on specific signals and are designed to be highly specialized. More importantly, they require high-temporal resolution data.

ML techniques have also been used to detect and remove artifacts ICU physiology data. One of the earliest ML-based work in artifact detection for ICU minute resolution time series data is Tsien \MakeLowercase{\textit{et al.}} \cite{Tsien2001}. This study formulated the task as a supervised learning problem using a decision tree model and a dataset with annotations by an expert. The time series were segmented into overlapping segments for which a feature vector was computed. Labels were assigned to each segment depending on whether more or less than 50\% of samples were artifacts. However, this approach potentially wastes data especially for long window sizes.

More recently, numerous ML techniques for artifact detection in ICU physiological time-series have been explored \cite{Mataczynski2022End-to-EndLearning,Subramanian2021UnsupervisedActivity, Chen2021SignalApproaches,Lee2020ArtifactInjury,Wu2020,Edinburgh2019}. Mataczynski \MakeLowercase{\textit{et al.}} \cite{Mataczynski2022End-to-EndLearning} used a deep learning model based on ResNets to classify Intracranial Pressure (ICP) waveform pulses as: normal, possibly pathological, likely pathological, pathological or artifact. Subramanian \MakeLowercase{\textit{et al.}} \cite{Subramanian2021UnsupervisedActivity} compared unsupervised ML models: $k$-nearest neighbours, isolation forests, and 1-class support vector machines (OCSVM) to detect artifacts in electrodermal activity data sampled at 256Hz. The time series were transformed into 0.5s windows and 12 features were extracted. The unsupervised models computed scores reflecting the presence of artifacts in the window. Chen \MakeLowercase{\textit{et al.}} \cite{Chen2021SignalApproaches} use a convolutional neural network (CNN) on 10s short-time Fourier transformed segments of a PPG signal to classify them as good or bad. Edinburgh \MakeLowercase{\textit{et al.}} \cite{Edinburgh2019} used an unsupervised model based on variational autoencoders (VAE) to detect artifacts in 10s windows of invasive arterial blood pressure. All of these techniques work on high-resolution data and detect artifacts at window/segment level. Hence, they are not directly applicable to routinely collected low-resolution clinical signals.

In contrast, Wu \MakeLowercase{\textit{et al.}} \cite{Wu2020} worked on relatively low-resolution invasive blood pressure time series with 1 sample every 15s and predicted the presence of artifacts at the sample level. The authors compared forecasting methods (ARIMA, exponential smoothing), XGBoost, isolation forest, and OCSVM, and found that XGBoost significantly outperformed other methods with over 99\% sensitivity and specificity. However, XGBoost being a fully-supervised model requires labeled data to train. In addition, the study showed that multivariate models which take in mean, systolic and diastolic blood pressure outperformed univariate models which take in systolic blood pressure only.

It is worth considering  techniques developed for detecting anomalies in other kinds of time series as well  since this problem closely resembles artifact detection in physiological time series. Munir \MakeLowercase{\textit{et al.}} \cite{Munir2019DeepAnT:Series} developed a CNN-based model with a predictor and anomaly detector module. The predictor module predicts the next timestamp sample, which is then fed into the anomaly detector module. The anomaly detector module computed the Euclidian distance between actual sample and predicted sample, which determined if a sample is an outlier based on a set threshold. The method could detect both point and contextual anomalies in periodic time series.

Numerous studies have used VAEs for artifact detection \cite{Edinburgh2019} and anomaly detection \cite{Lin2020AnomalyModel,Niu2020LSTM-basedDetection,Gangloff2022LeveragingDetection}, including a $\beta$-VAE \cite{Cheng2023VariationalTagging}. However, all of these studies train their VAEs with data that do \textit{not} contain anomalies, making all these approaches ``weakly'' unsupervised since labels are indeed required to filter the data prior to training, as remarked by Gangloff \MakeLowercase{\textit{et al.}} \cite{Gangloff2022LeveragingDetection}. In contrast, our proposed approach is ``completely'' unsupervised, since we train our VAE on unfiltered data. For training their models, Lin \MakeLowercase{\textit{et al.}} and Niu \MakeLowercase{\textit{et al.}} split a time series recording into two parts: the first part for training and the last for testing. This strategy requires a model to have seen part of the time series and is less practical for detecting artifacts/anomalies on retrospective data. Our approach splits whole recordings into training and testing sets, ensuring no information leakage from the training to the test set.

\section{Methodology}\label{methodology}

In this work, we develop a completely unsupervised approach utilizing $\beta$-VAE \cite{Higgins2017} and IF \cite{Liu2008IsolationForest} to detect artifacts at the sample level in minute-resolution vital signs data, including mean blood pressure (BPm), heart rate (HRT), and mean intracranial pressure (ICPm). This approach aims to eliminate the need for labeled data when building models for artifact detection. Unlike previous studies utilizing VAEs to detect artifacts/anomalies by first training on clean data \cite{Cheng2023VariationalTagging,Edinburgh2019,Lin2020AnomalyModel,Niu2020LSTM-basedDetection,Gangloff2022LeveragingDetection}, our method trains on raw data containing both clean samples and artifacts. We show that the model learns and generalizes the dynamics of artifacts across different vital signs by training our model on a dataset of all the mentioned vital signs mixed. Additionally, we develop a fully supervised classifier based on LSTM with attention as a proposed benchmark \cite{Hochreiter1997, Vaswani2017AttentionNeed}, alongside implementing Isolation Forest on its own, XGBoost and ARIMA as additional benchmarks. An ablation study demonstrates the significance of feature extraction with a VAE and the importance of the self-attention component in the LSTM classifier, including training an LSTM classifier without the self-attention component to highlight its significance. We also explore the latent space of real-world data containing artifacts. We validate our method using gold-standard annotations by an experienced expert and demonstrate its generalizability on the MIMIC-IV \cite{MoodyMIMIC-IVDatabase} dataset without retraining. As highlighted in Section~\ref{introduction}, the scarcity of labeled data makes our approach particularly valuable. While recognizing this scarcity, we first establish a performance benchmark using supervised models before presenting and evaluating our novel unsupervised approach, showcasing its generalizability to an external dataset. Before delving into the details of the models, we describe the dataset, preprocessing, and problem formulation.

\subsection{KidsBrainIT dataset}
\begin{figure}[!t]
\centerline{\includegraphics[width=\columnwidth]{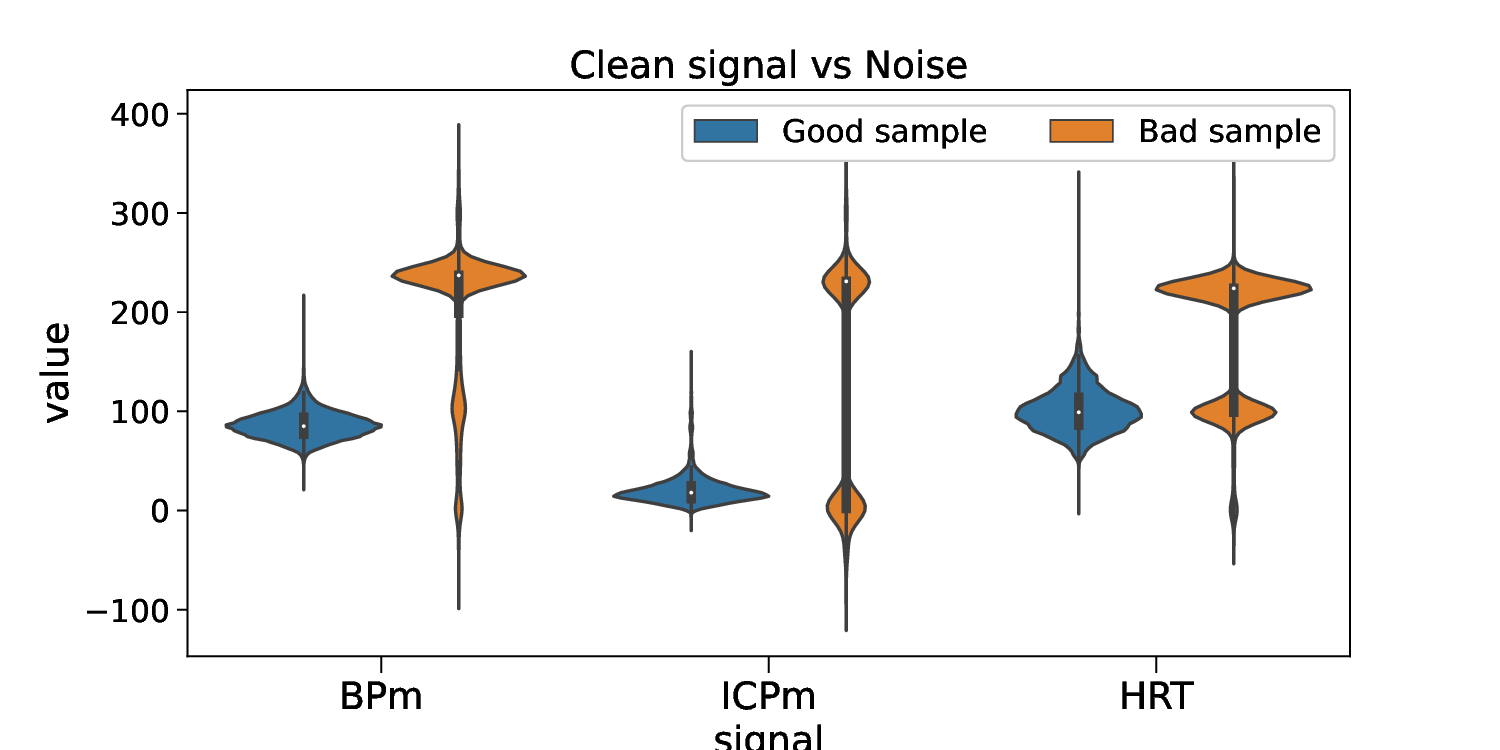}}
\caption{Distribution of amplitude values for artifactual and non-artifactual samples in KidsBrainIT dataset.}
\label{artifacts_dist}
\end{figure}

We have access to a research dataset from the KidsBrainIT project \cite{Lo2018} that was collected prospectively and extracted for research purposes from Paeditric Intensive Care Units (PICU) with approval reference number 17/WS/0086 by the Research Ethics Committee in the United Kingdom. We use the KidsBrainIT pilot dataset, which contains about 98 patients from 2 PICU in the UK and about 27 patients that were added later on, making up a total of 125. Each patient in the  dataset has around 14 physiological signals that were collected at minute-by-minute intervals, a routine clinical setup in ICU. It also includes annotations on artifacts by an expert researcher. For our experiments, we used the BPm, ICPm, and HRT as these signals have the least amount of missing values across all patients recordings. Fig.~\ref{artifacts_dist} shows the distribution of artifacts and good signal for the selected vital signs.

\subsection{Data preprocessing}\label{sec:preproc}
\begin{figure}[!t]
\centerline{\includegraphics[width=\columnwidth]{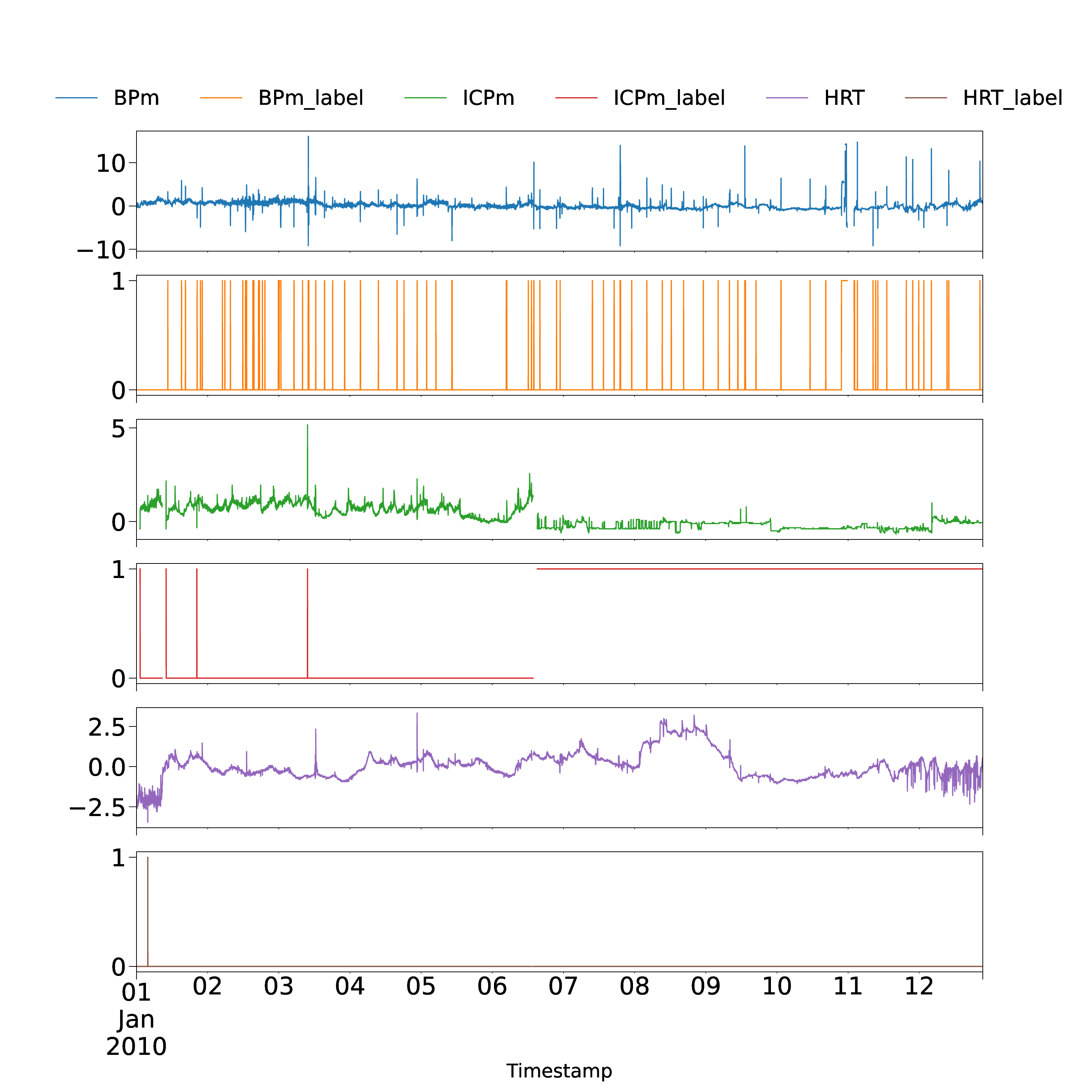}}
\caption{Sample patient standardised recording, containing three signals (BPm, ICPm, HRT) and their corresponding artifact labels,  (BPm\_label, ICP\_label, HRT\_label), as annotated by an expert researcher.}
\label{sample_recording}
\end{figure}

\begin{figure}[!t]
\centerline{\includegraphics[width=\columnwidth]{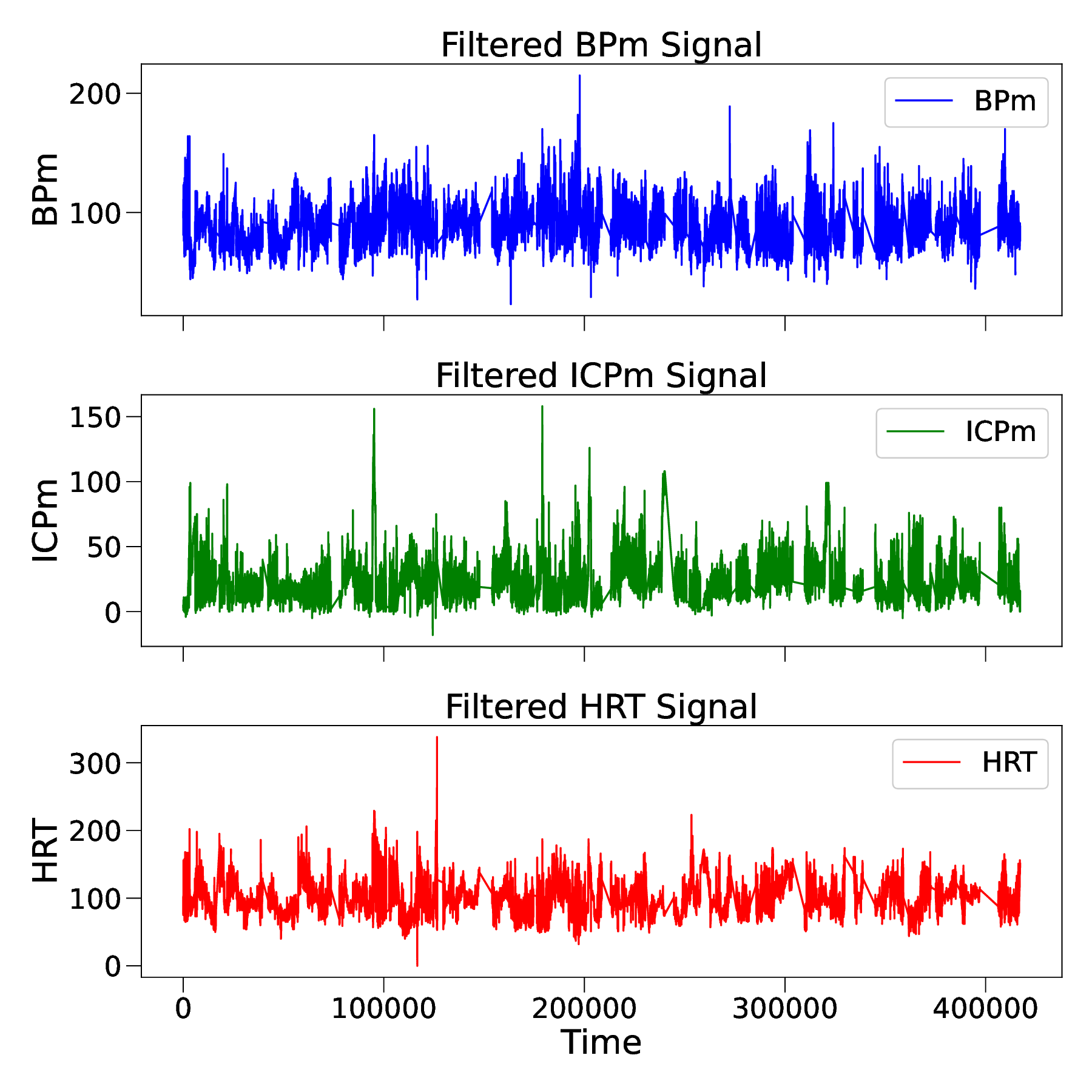}}
\caption{Figure showing concatenated signals across all patients showing that the three signals are non-periodic}
\label{fig:periodicity}
\end{figure}

We selected 64 patients out of the 125, with at least 90\% data in BPm, ICPm, and HRT recordings. In Fig.~\ref{fig:periodicity} we plot the concatenated and filtered signals i.e. artifacts removed, showing that the three signals do not exhibit any periodic behavior. Table~\ref{tab:signal_stats}, shows statistics and correlation matrix, showing weak or no dependency between BPm, ICPm, and HRT. We then randomly split these patients into 47 and 17 as training and testing sets respectively. We standardised all patient recordings individually by subtracting the median followed by division with the interquartile range of the recording. Standardizing each recording individually ensures that the differences in vital sign dynamics between patients are preserved. We also applied quadratic interpolation to fill in the missing values in the standardized recordings. We created a sequence of sample labels from each patient recording based on annotations by the expert, i.e. 1 for ``artifact'' and 0 otherwise. We use these labels as ground truth to measure the performance of different models. To preserve privacy, we shifted the actual timestamps of the recordings to midnight \nth{1} Jan, 2010. Fig.~\ref{sample_recording} shows an example of a pre-processed patient recording.  We then applied an overlapping sliding window of size $W$ minutes, and step size of 1 minute (1 sample) on the recordings across all patients in the training set. $W$ is a hyperparameter that is tuned during the experiments. 

\begin{table}[h!]

\centering
\begin{tabular}{lrrrrr}
\toprule
{} &       Mean &    Std Dev &   Variance &    Min &    Max \\
\midrule
BPm  &  85.53 &  12.46 & 155.28 &  23.0 &  215.0 \\
ICPm &  20.30 &  13.77 & 189.65 & -18.0 &  158.0 \\
HRT  &  98.35 &  20.86 & 435.07 &   0.0 &  338.0 \\
\bottomrule
\end{tabular}
\vspace{0.5cm}
\begin{tabular}{lrrr}
\toprule
{} &       BPm &      ICPm &       HRT \\
\midrule
BPm  &  1.00 &  0.21 & -0.04 \\
ICPm &  0.21 &  1.00 &  0.06 \\
HRT  & -0.04 &  0.06 &  1.00 \\
\bottomrule
\end{tabular}
\caption{Statistical properties and correlation matrix showing weak or no dependency between BPm, ICPm and HRT}
\label{tab:signal_stats}
\end{table}

\subsection{Problem formulation}\label{sec:propform}
We consider a preprocessed dataset of size $N$, $D:=  \left \{ \left ( \textbf{x}_{i},\textbf{y}_{i} \right ) \right \}$ such that $\textbf{x} \in \mathbb{R}^{W}, \textbf{y}\in \left \{ 0,1 \right \}^{W},\mathit{i}\in \left \{ 1 \ldots N  \right \}$, where $\textbf{x}_{i}$ is a segment of length $W$ of the time series and $\textbf{y}_{i}$ is the corresponding label for each \textit{sample} in the segment as described in Section~\ref{sec:preproc}. Additionally, we define $\textbf{z} \in \mathbb{R}^{H}$, a vector of size $H$, as the latent representation of a sample $x$ in the segment $\textbf{x}$. The latent representation for a signal of length $L$ is represented by a matrix $\textbf{Z}\in \mathbb{R}^{L \times H}$.

\subsection{Fully unsupervised approach: VAE-IF}
In this section, we present our proposed unsupervised approach for artifact detection in minute-by-minute physiological recordings. Notably, this approach is particularly valuable as it does not rely on labeled data. At a high level, our approach involves pre-training a VAE \cite{Kingma2014} model on the preprocessed training dataset, which includes both noisy and clean segments, $\mathbf{x}_i$, as described in Sections~\ref{sec:preproc} and \ref{sec:propform}. Subsequently, we utilize the encoder of the trained VAE to extract features from a test signal. Once the features are extracted, we directly employ the IF \cite{Liu2008IsolationForest} anomaly detection algorithm to classify samples in the signal as normal or artifact (anomaly). In the following subsections, we give a detailed description of the main components of this approach.

\subsubsection{Training the VAE}
\begin{figure}[!t]
  \centering
  \includegraphics[width=0.35\linewidth]{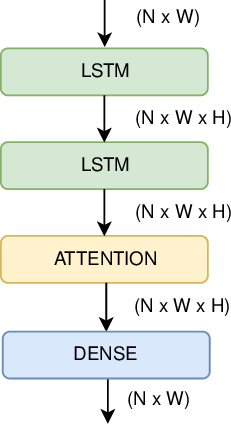}
  \caption{Architectual diagram of the supervised LSTM-Attn benchmark.}
  \label{fig:lstm_attn}
\end{figure}

\begin{figure}[!t]
  \centering
  \includegraphics[width=0.5\linewidth]{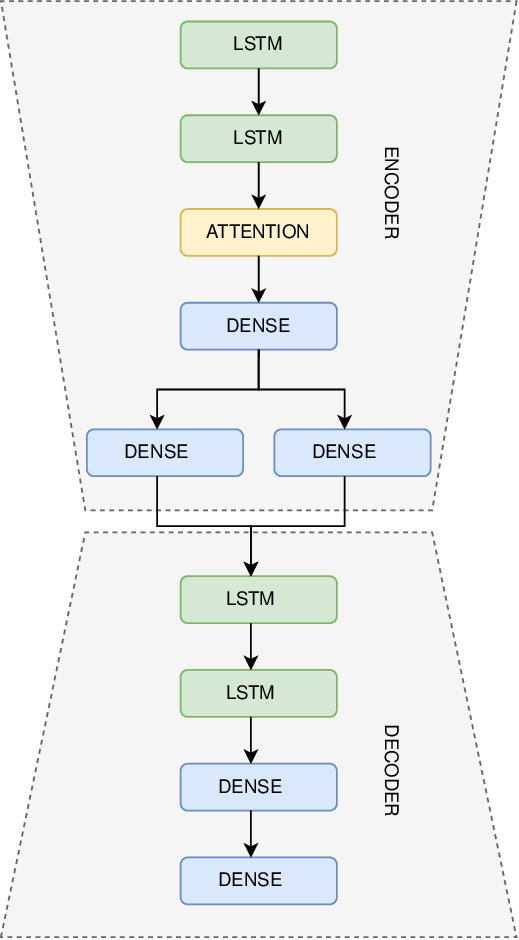}
  \caption{Architectural diagram of the VAE.}
  \label{fig:vae}
\end{figure}

Our model is based on $\beta$-VAE \cite{Higgins2017}, a variant of VAE with the following loss function (Evidence Lower Bound, ELBO) in Eq.~\ref{eqn:ELBO_reg}:
\begin{equation}
\label{eqn:ELBO_reg}
\mathcal{L}(\theta ,\phi ; \mathbf{x}) = \mathbb{E}_{q_\phi (\mathbf{z|x})}\left [ \log p_\theta (\mathbf{x|z}) \right ] - \beta D_{KL}(q_\phi (\mathbf{z|x})||p_\theta (\mathbf{z})),
\end{equation}
where $\theta, \phi$ are the parameters of the decoder $p(\mathbf{x}|\mathbf{z})$ and encoder $q(\mathbf{z}|\mathbf{x})$ and $\beta$ is a regularization coefficient. Tuning $\beta$ has a smoothing effect on the high-frequency components of the signals, enabling the model to learn stable representations, which is our main motivation for using this variant of VAEs. The encoder consists of two stacked LSTM layers, followed by attention and a dense layer. Similarly, the decoder has a comparable architecture, except for the absence of the attention block. The first three layers of the encoder are similar to that of the LSTM-Attn benchmark model in Fig~\ref{fig:lstm_attn}. The operation of these LSTM blocks is described in Section~\ref{sec:LSTM}. The full architecture for the VAE can be seen in Fig.~\ref{fig:vae}.

We train the VAE with a mix of clean and artifact-containing data segments $\textbf{x}$. We use Bayesian optimization (BO) \cite{Mockus1978TheExtremum} to tune model hyper-parameters listed in Table~\ref{tab:hyperparameters_vae}.

\subsubsection{Feature extraction and averaging}
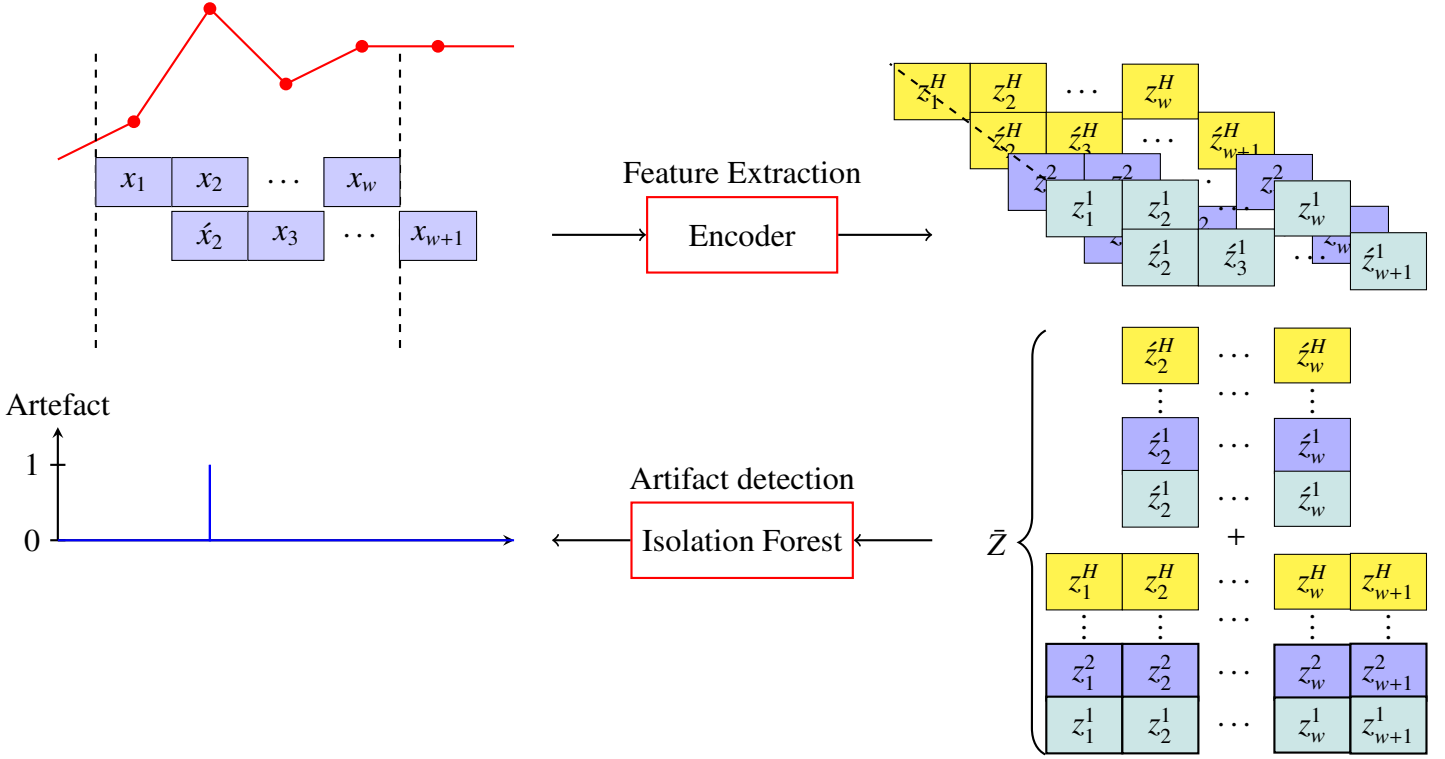
\begin{figure*}
  \centering
  \begin{tikzpicture}[
    thick, text centered,
    box/.style={draw, thin, minimum width=1cm, minimum height=0.65cm},
    box-hidden/.style={draw=none, thin, minimum width=1cm, minimum height=0.65cm},
    hbox/.style={draw=red, minimum width=2.5cm, minimum height=1cm, align=center},
    func/.style={circle, text=white},
    input/.style={draw=black},
    vector/.style={draw, minimum width=1cm, minimum height=0.65cm}
  ]
  \draw[white] (-0.5,-2.5) rectangle (18.5,2.5);

  
  \draw[red] (0,1) -- (1,1.5) -- (2,3) -- (3,2) -- (4,2.5) -- (5,2.5) -- (6,2.5)
  plot[only marks,mark=*] coordinates{(1,1.5) (2,3) (3,2) (4,2.5) (5,2.5)};

  \draw[dashed] (0.5,-1.5) -- (0.5,2.5);
  \draw[dashed] (4.5,-1.5) -- (4.5,2.5);

  \node[box, input, fill=blue!20, shift={(1,0.7)}] (x1) {$x_1$};
  \node[box, input, fill=blue!20, right of=x1] (x2) {$x_2$};
  \node[right of=x2] (xdots1) {\dots};
  \node[box, input, fill=blue!20, right of=xdots1] (xw) {$x_w$};

  \node[box, fill=blue!20, below=0.05 of x2] (z1) {$\acute{x}_2$};
  \node[box, fill=blue!20, right of=z1] (z2) {$x_3$};
  \node[right of=z2] (zdots1) {\dots};
  \node[box, fill=blue!20, right of=zdots1] (zw) {${x}_{w+1}$};



  \node[box-hidden, fill=white, fill opacity=0.0,shift={(8.5,2.6)}] (TN) at (2,0) {};
  \begin{scope}[shift={(11.5,1.9)}]
    \node[box, input, fill=yellow!80] (h1H) at (0,0) {$z_{1}^{H}$};
    \node[box, input, fill=yellow!80] (h2H) at (1,0) {$z_{2}^{H}$};
    \node[] (hdash) at (2,0) {\dots};
    \node[box, input, fill=yellow!80] (hwH) at (3,0) {$z_{w}^{H}$};
  \end{scope}
  \begin{scope}[shift={(12.5,1.25)}]
    \node[box, input, fill=yellow!80] (hh1H) at (0,0) {$\acute{z}_{2}^{H}$};
    \node[box, input, fill=yellow!80] (hh2H) at (1,0) {$\acute{z}_{3}^{H}$};
    \node[] (hdash) at (2,0) {\dots};
    \node[box, input, fill=yellow!80] (hhwH) at (3,0) {$\acute{z}_{w+1}^{H}$};
  \end{scope}

  \begin{scope}[shift={(13,0.7)}]
    \node[box, input, fill=blue!30] (h12) at (0,0) {$z_{1}^{2}$};
    \node[box, input, fill=blue!30] (h22) at (1,0) {$z_{2}^{2}$};
    \node[] (hdash) at (2,0) {\dots};
    \node[box, input, fill=blue!30] (hw2) at (3,0) {$z_{w}^{2}$};
  \end{scope}
  \begin{scope}[shift={(14,0)}]
    \node[box, input, fill=blue!30] (hh12) at (0,0) {$\acute{z}_{2}^{2}$};
    \node[box, input, fill=blue!30] (hh22) at (1,0) {$\acute{z}_{3}^{2}$};
    \node[] (hdash) at (2,0) {\dots};
    \node[box, input, fill=blue!30] (hhw2) at (3,0) {$\acute{z}_{w+1}^{2}$};
  \end{scope}
    \begin{scope}[shift={(13.5,0.35)}]
    \node[box, input, fill=teal!20] (h11) at (0,0) {$z_{1}^{1}$};
    \node[box, input, fill=teal!20] (h21) at (1,0) {$z_{2}^{1}$};
    \node[] (hdash) at (2,0) {\dots};
    \node[box, input, fill=teal!20] (hw1) at (3,0) {$z_{w}^{1}$};
  \end{scope}
    \begin{scope}[shift={(14.5,-0.3)}]
    \node[box, input, fill=teal!20] (hh11) at (0,0) {$\acute{z}_{2}^{1}$};
    \node[box, input, fill=teal!20] (hh21) at (1,0) {$\acute{z}_{3}^{1}$};
    \node[] (hdash) at (2,0) {\dots};
    \node[box, input, fill=teal!20] (hhw1) at (3,-0.05) {$\acute{z}_{w+1}^{1}$};
    \draw[dashed](h11)--(TN);
  \end{scope}

\begin{scope}[shift={(9,0)}]
      
      \node[hbox] (encoder) {Encoder};
      \node[above=15pt]{Feature Extraction};
    
      \draw[->] (-2.5,0) -- (encoder.west) node[midway, above] {};
      \draw[->] (encoder.east) -- (2.5,0) node[midway, above] {};
  \end{scope}

\begin{scope}[shift={(14.5,-1.6)}]
    \node[box-hidden, fill=white] (fstart) at (-1,0) {}; 
    \node[box, input, fill=yellow!80] (hh11) at (0,0) {$\acute{z}_{2}^{H}$};
    \node[] (hdash) at (1,0) {\dots};
    \node[box, input, fill=yellow!80] (hh21) at (2,0) {$\acute{z}_{w}^{H}$};
  \end{scope}
\begin{scope}[shift={(14.5,-2.1)}]
    \node[] (hh11) at (0,0) {\vdots};
    \node[] (hdash) at (1,0) {\dots};
    \node[] (hh11) at (2,0) {\vdots};
  \end{scope}
\begin{scope}[shift={(14.5,-2.8)}]
    \node[box, input, fill=blue!30] (hh11) at (0,0) {$\acute{z}_{2}^{1}$};
    \node[] (hdash) at (1,0) {\dots};
    \node[box, input, fill=blue!30] (hh21) at (2,0) {$\acute{z}_{w}^{1}$};
  \end{scope}
\begin{scope}[shift={(14.5,-3.5)}]
    \node[box, input, fill=teal!20] (hh11) at (0,0) {$\acute{z}_{2}^{1}$};
    \node[] (hdash) at (1,0) {\dots};
    \node[box, input, fill=teal!20] (hh21) at (2,0) {$\acute{z}_{w}^{1}$};
  \end{scope}
 \begin{scope}[shift={(13.5,-4.0)}]
    \node[] (h11) at (0,0) {};
    \node[] (h11) at (1,0) {};
    \node[] (hdash) at (2,0) {+};
    \node[] (h11) at (3,0) {};
    \node[] (h11) at (4,0) {};
  \end{scope}
 \begin{scope}[shift={(13.5,-4.6)}]
    \node[box, input, fill=yellow!80] (h1H) at (0,0) {$z_{1}^{H}$};
    \node[box, input, fill=yellow!80] (h2H) at (1,0) {$z_{2}^{H}$};
    \node[] (hdash) at (2,0) {\dots};
    \node[box, input, fill=yellow!80] (hwH) at (3,0) {$z_{w}^{H}$};
    \node[box, input, fill=yellow!80] (hwH) at (4,0) {$z_{w+1}^{H}$};
  \end{scope}
\begin{scope}[shift={(13.5,-5.1)}]
    \node[] (h11) at (0,0) {\vdots};
    \node[] (h11) at (1,0) {\vdots};
    \node[] (hdash) at (2,0) {\dots};
    \node[] (h11) at (3,0) {\vdots};
    \node[] (h11) at (4,0) {\vdots};
  \end{scope}
\begin{scope}[shift={(13.5,-5.8)}]
    \node[vector, fill=blue!30] (h11) at (0,0) {$z_{1}^{2}$};
    \node[vector, fill=blue!30] (h21) at (1,0) {$z_{2}^{2}$};
    \node[] (hdash) at (2,0) {\dots};
    \node[vector, fill=blue!30] (hw1) at (3,0) {$z_{w}^{2}$};
    \node[vector, fill=blue!30] (hw1) at (4,0) {$z_{w+1}^{2}$};
  \end{scope}
\begin{scope}[shift={(13.5,-6.5)}]
    \node[vector, fill=teal!20] (fend) at (0,0) {$z_{1}^{1}$};
    \node[vector, fill=teal!20] (h21) at (1,0) {$z_{2}^{1}$};
    \node[] (hdash) at (2,0) {\dots};
    \node[vector, fill=teal!20] (hw1) at (3,0) {$z_{w}^{1}$};
    \node[vector, fill=teal!20] (hw1) at (4,0) {$z_{w+1}^{1}$};
  \end{scope}

  \draw[decorate, decoration={brace, amplitude=10pt, mirror}] (fstart.north west) -- (fend.south west) node[midway,left=10pt] {$\bar{Z}$};

  \begin{scope}[shift={(9,-4.05)}]
      
      \node[hbox] (IF) {Isolation Forest};
      \node[above=15pt]{Artifact detection};
    
      \draw[->] (2.5,0) -- (IF.east) node[midway, above] {};
      \draw[->] (IF.west) -- (-2.5,0) node[midway, above] {};
  \end{scope}

\begin{scope}[shift={(0,-4.05)}]
\draw[-stealth] (0,0) -- (0,1.5) node[above] {Artefact};
\draw[-stealth] (0,0) -- (6,0) node[right] {};

\draw (-0.1,0) -- (0.1,0) node[left, xshift=-5pt, yshift=-0pt] {0};
\draw (-0.1,1) -- (0.1,1) node[left, xshift=-5pt, yshift=0pt] {1};
  \draw[blue, thick ] (0,0) -- (1,0) -- (2,0) -- (2,1) -- (2,0) -- (3,0) -- (4,0) -- (4,0) -- (5,0) -- (5,0) -- (6,0);  
\end{scope}

\end{tikzpicture}
  \caption{Feature extraction via overlapping sliding-window on target signal \(x\). This process employs a trained VAE for encoding, leading to multiple latent representations (e.g., \(z\) and \(z'\)) for some signal samples based on their position. These representations are averaged to produce a consistent signal representation, \(\bar{Z}\), which is subsequently input to the iForest model for artifact detection.
}
  \label{fig:methodology}
\end{figure*}

We use the encoder of a trained VAE to extract features from a target physiological time series. We apply a sliding window with unit step size on the target signal, resulting in overlapping segments. Passing these overlapping segments through the encoder to extract features will result in some samples having more than one embedding vector. This is crucial since averaging these embeddings gives a more accurate representation of the sample. This operation is summarized in Fig.~$\ref{fig:methodology}$. Essentially, each sample in the original signal is converted into a vector that represents the state of the sample given the previous samples. The transformed sample windows are then averaged to obtain a sequence of state vectors $\bar{Z}$.

\subsubsection{Artifact detection with IF}
IF \cite{Liu2008IsolationForest} is an algorithm for anomaly detection which uses an ensemble of isolation trees to partition data based on the assumption that anomalies exhibit distinct characteristics that set them apart from normal data points. Consequently, anomalies are expected to be isolated closer to the root node of the trees compared to normal data points. Compared to other algorithms like OCSVM, IF is computationally faster and scales well to high-dimensional datasets \cite{Liu2008IsolationForest}.

We use Sklearn's \cite{Pedregosa2011Scikit-learn:Python} implementation of IF with its default hyperparameters for artifact detection. IF produce a sequence with output values corresponding to clean and artifactual samples, represented by the vector sequence $\bar{Z}$. We first train the IF model on the extracted features of all patients and the corresponding target signals in training set before using it on the testing set.

\begin{table}[htbp]
\centering
\caption{Hyperparameters and ranges included in the BO.}
\label{tab:hyperparameters_combined}

\begin{subfigure}[t]{\linewidth}
    \centering
    \caption{VAE-IF}
    \label{tab:hyperparameters_vae}
    \begin{tabular}{ll}
        \hline
        \textbf{Hyperparameter} & \textbf{Range} \\
        \hline
        lr & [1e-4, 0.001] (log scale) \\
        latent\_dim & [2, 90] \\
        $\beta$ & [1e-3, 1e2] (log scale) \\
        n\_epochs & [8, 30] \\
        window\_size & [15, 60] \\
        bidirectional & \{True, False\} \\
        \hline
    \end{tabular}
\end{subfigure}

\vspace{0.5cm}

\begin{subfigure}[t]{\linewidth}
    \centering
    \caption{LSTM-ATTn}
    \label{tab:hyperparameters_lstm}
    \begin{tabular}{ll}
        \hline
        \textbf{Hyperparameter} & \textbf{Range} \\
        \hline
        lr & [1e-4, 0.001] (log scale) \\
        n\_epochs & [8, 30] \\
        window\_size & [15, 60] \\
        bidirectional & \{True, False\} \\
        \hline
    \end{tabular}
\end{subfigure}

\vspace{0.5cm}

\begin{subfigure}[t]{\linewidth}
    \centering
    \caption{XGBoost}
    \label{tab:hyperparameters_XGBoost}
    \begin{tabular}{ll}
        \hline
        \textbf{Hyperparameter} & \textbf{Range} \\
        \hline
        lr & [0.01, 0.5]  \\
        n\_estimators & [100, 1000] \\
        window\_size & [15, 60] \\
        \hline
    \end{tabular}
\end{subfigure}

\end{table}

\subsection{Proposed benchmarks}
In this section, we describe the supervised models to be used as a performance benchmarks for the unsupervised approach. For this task, we: build a neural-network model based on LSTM with self-attention, and also use XGBoost and ARIMA, a standard statistical model.

\subsubsection{LSTM with self-attention (LSTM-ATTn)}
\label{sec:LSTM}
LSTM \cite{Hochreiter1997} is a type of recurrent neural network architecture for learning sequential data. An LSTM can learn temporal information over long sequences through a hidden state. The hidden state at a time step is given by the following equation,
\begin{equation}
\label{LSTM}
\mathbf{h}_t = LSTM(x_t,\mathbf{h}_{t-1}),
\end{equation}
where $\mathbf{h}_t \in \mathbb{R}^{H}$ is the hidden state vector and $x_i \in \mathbb{R}$ is a sample in segment $\mathbf{x}$ as defined in \ref{sec:propform}. A Bidirectional LSTM \cite{Schuster1997BidirectionalNetworks} has access to past and future information. Its hidden state has an effective size of $\mathbf{h}_t \in \mathbb{R}^{2H}$.

We stack two LSTM layers with hidden size $H$ followed by a scaled dot-product attention \cite{Vaswani2017AttentionNeed}, given by
\begin{equation}
    \label{attention}
    {\mathbf{h}}'_i = \sum_{j}^{W}\mathrm{softmax}\left(\frac{\mathbf{h}_i^{T}\mathbf{h}_j}{\sqrt{H}}\right)\mathbf{h}_j,
\end{equation}
where ${\mathbf{h}}'_i$ is the final hidden state and $W$ is the window size as defined in Section~\ref{sec:preproc}. We also explore the effect of using a bidirectional LSTM during hyper-parameter tuning. Next, we add a dense layer with a sigmoid activation function to give a probability estimate ${y}'_i$ of sample $x_i$ being an artifact. We train the model using Adam optimizer \cite{Kingma2015Adam:Optimization}. Fig~\ref{fig:lstm_attn} shows the architecture diagram of the LSTM-Attn classifier.



\subsubsection{XGBoost}
XGBoost \cite{Chen2016XGBoost:System}, or Extreme Gradient Boosting, is a decision tree ensemble based on CART (Classification and Regression Trees)  models trained using the gradient boosting algorithm \cite{Friedman2001GreedyMachine}. The algorithm trains additively by fitting new trees to the residuals. XGBoost is known for its excellent performance \cite{Wu2020}, and we use it as a second benchmark. We use XGBoost to classify each part of a signal as either an artifact or not. The input is the original signal, and the output is a label for each sample in that signal.

\subsubsection{ARIMA}
In addition to ML-based models, we benchmark using an ARIMA \cite{Geurts1977TimeControl} model, which is a popular statistical model for time series analysis. Autoregressive Integrated Moving Average (ARIMA) is a statistical model for representing a stochastic non-stationary signal, assuming the signal has stationary and non-stationary components. The non-stationary component is represented by stochastic autoregressive (AR) process of order $P$ (i.e. the number of steps in the past required to predict the current value). The stationary component is represented by a moving average (MA) process of order $Q$ (i.e. the number of neighboring steps to average). To make the signal stationary, differencing operations of order $D$ are applied to the signal. We fit a signal with ARIMA model and treat residuals more than three standard deviations from the mean as artifacts.

\subsection{Ablation study}
We also perform an ablation study to investigate the individual contributions and significance of various components in our proposed methods. By systematically removing specific elements and evaluating their impact on the performance, we gain insights into the effectiveness of each component.

\subsubsection{Feature extraction ablation}
To assess the importance of feature extraction using $\beta$-VAE \cite{Higgins2017}, we compare the performance of model with a vanilla Isolation Forest model trained directly on the patient recordings. By excluding the VAE-based feature extraction step, we examine whether the extracted features significantly contribute to the detection of artifacts in minute-resolution vital signs data.

\subsubsection{Self-Attention Ablation}
We also analyze the significance of the self-attention component in the encoder of the VAE. We exclude the self-attention component in the encoder of the VAE and train the model with the same set of hyper-parameters estimated by BO. This ablation study helps us understand the added benefit of incorporating self-attention for accurate artifact detection.

\section{Experiments and Results}\label{results}
We used BO to tune model hyperparameters for the VAE part of the VAE-IF, LSTM-ATTn, and XGBoost, using Geometric Mean as the objective function. We also explored training hyperparameters such as window size ($W$), learning rate, and number of epochs. We use the function ``\textit{auto\_arima}'' from the package \textit{pmdarima}\cite{pmdarima} to find the best ARIMA model (i.e., $P$, $Q$, $D$) for each of the three signals in the training set. The range of hyperparameters explored for the other models is summarised in Table~\ref{tab:hyperparameters_combined}.

Using the test set, we evaluate the performance of VAE-IF and compare it to ARIMA and the supervised machine models: LSTM-ATTn and XGBoost. To measure performance, we compute sensitivity and specificity on a \textit{sample-by-sample} basis. We compute the performance metrics from each recording in the test set separately and report the mean and standard deviation scored by each model.

We visualize the latent space learned by the VAE using t-SNE \cite{VanDerMaaten2008VisualizingT-SNE} to show how clean signals and artifactual signals are represented in the latent space. We use default parameters for t-SNE, with the exception of perplexity which we set to 5. 

We use Pytorch and Pyro \cite{Bingham2019Pyro:Programming} to implement the VAE. BO was performed using \textit{ax.dev} \cite{Bakshy2018AE:Experimentation}, an open-source platform for managing adaptive experiments. We trained our models on Nvidia's GeForce RTX 3080 GPUs.

\subsection{Unsupervised approach}
We performed BO to tune hyperparameters listed in Table~\ref{tab:hyperparameters_vae}. Tuning hyperparameters was done in 60 iterations, whereby in each iteration, 70$\%$ of the training set was used for training and 30$\%$ for validation. The best-performing hyperparameters were then used to train the VAE and the final evaluation was done on the test set. The best-performing hyperparameters can be seen in the Supplementary materials. Table~\ref{tab:combined_performance} shows the performance results of VAE-IF on the test set. The table shows that VAE-IF achieves the highest sensitivity in two of the three signals.

\subsection{Benchmarks}
Similarly, we tune the hyperparameters for the LSTM-ATTn and XGBoost using BO. The best hyperparameters for the benchmark models also included in the supplementary material. Table~\ref{tab:combined_performance} show the best performance on test set for LSTM-ATTn, XGBoost, and ARIMA. Results show that LSTM-ATTn achieve the highest sensitivity for HRT signal. The benchmark models achieved higher specificity compared to VAE-IF across all three signals.

\subsection{External validation}
\begin{figure}[!t]
\centerline{\includegraphics[width=\columnwidth]{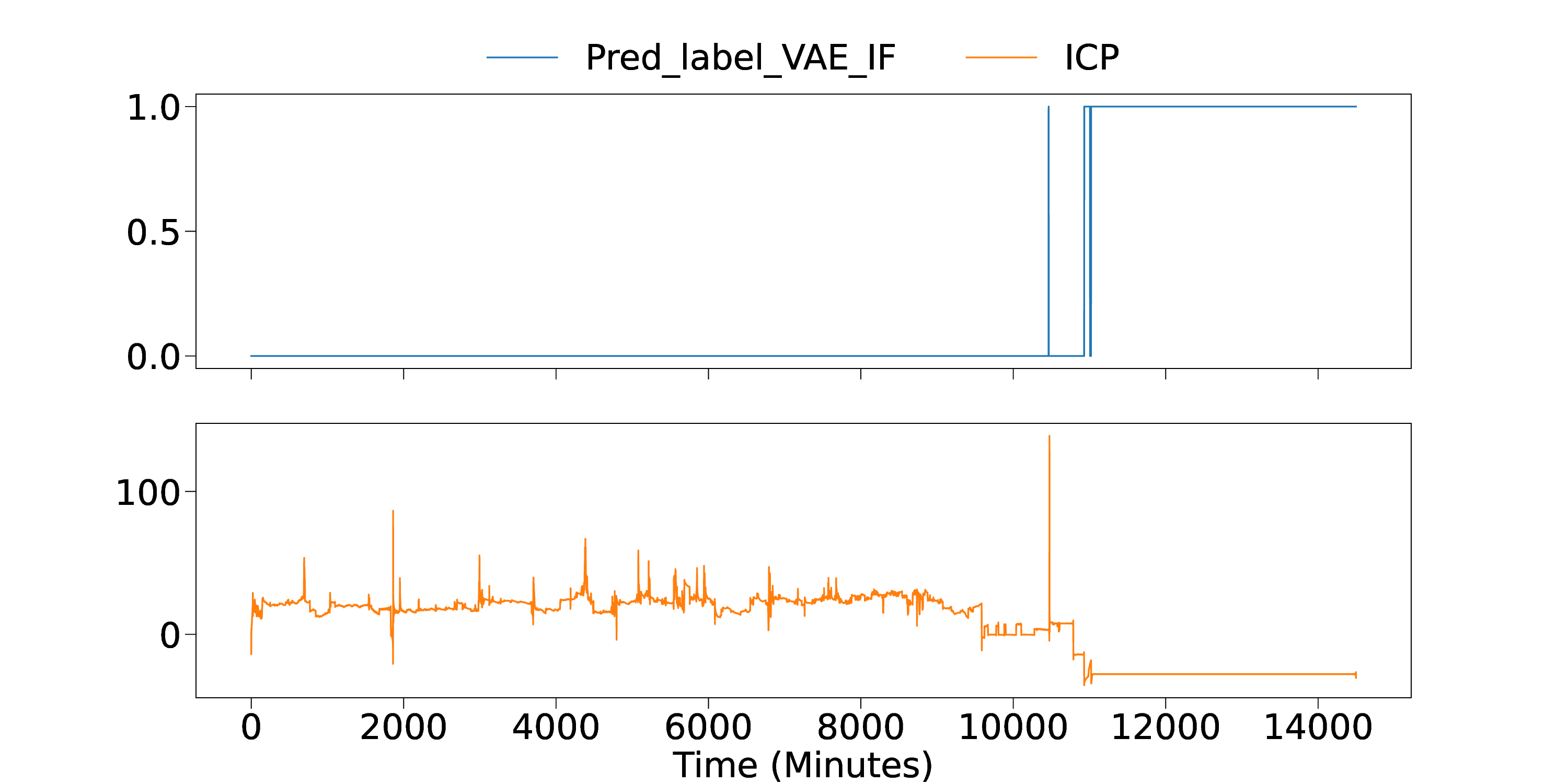}}
\caption{VAE\_IF prediction on patient p15649186 from MIMIC-IV waveform dataset, showing automatic detection of spikes and flatline artifacts without any retraining.}
\label{external_validation}
\end{figure}

We tested the generalizability of VAE-IF model on sample patients from the MIMIC-IV \cite{MoodyMIMIC-IVDatabase} waveform dataset. Here, we present results for the ICP signal of patient \textit{p15649186}. Given the difference in sampling rates between the datasets, we downsampled the ICP signal from 250Hz to 0.0167Hz (one sample per minute). We then applied the VAE-IF \textit{without any retraining}. Fig.~\ref{external_validation} shows the model's artifact prediction and the patient recording. Additional examples from more than 60 other subjects are included in the supplementary material.

\subsection{Latent space visualization with t-SNE}
\begin{figure}[!t]
\centerline{\includegraphics[width=\columnwidth]{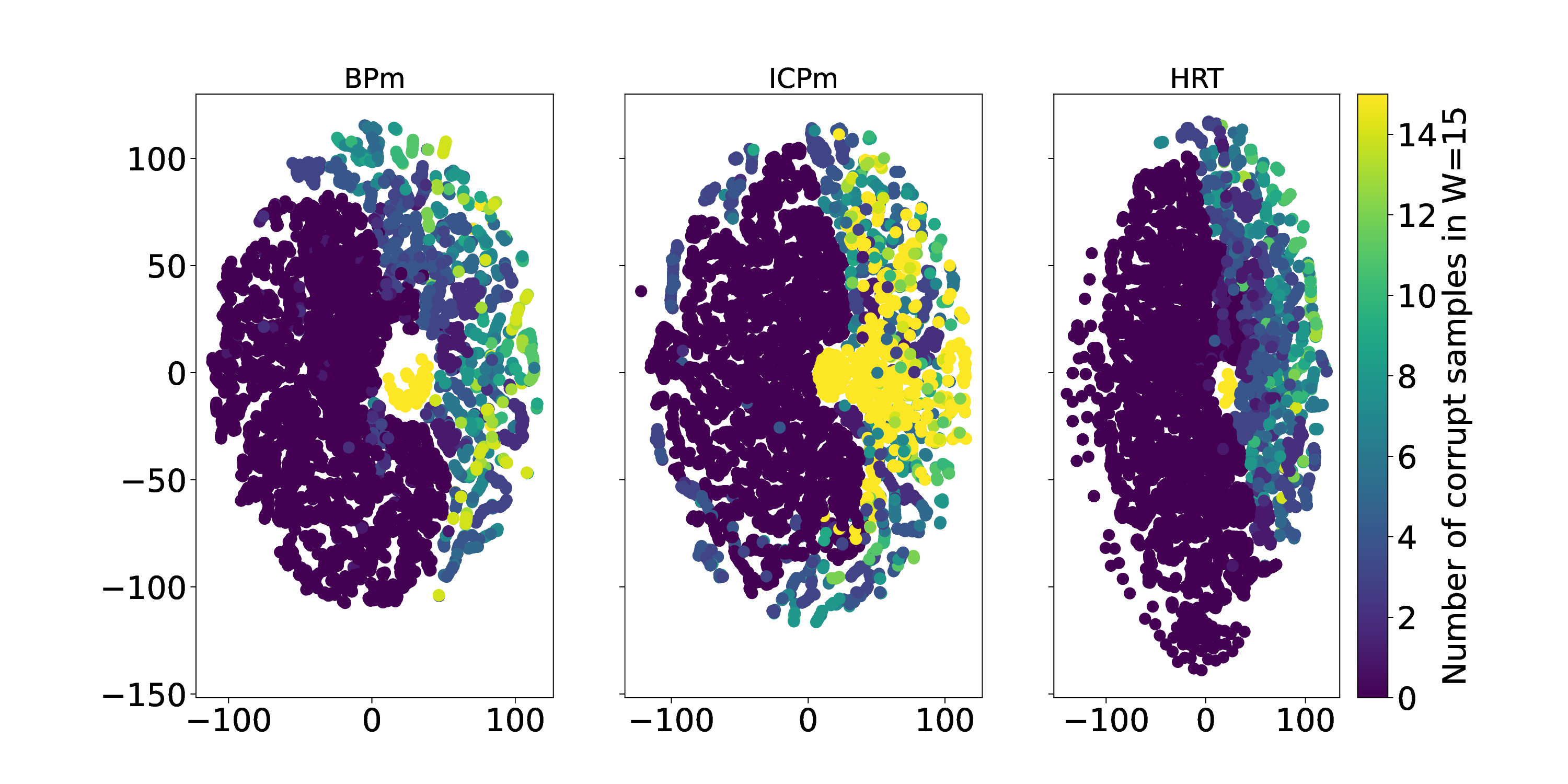}}
\caption{t-SNE visualization of a sample recording, showing disentangled latent representation of clean data from artifacts.}
\label{latent_space}
\end{figure}

We performed an exploration of the latent space embeddings $\textbf{z}$ based on one of the patients in KidsBrainIT (N004) by projection to 2D with t-SNE. Using $W = 15$, we first extract embeddings as illustrated in Fig.~\ref{fig:methodology}. Each embedding $\textbf{z}_i$ is given a numeric label of value between 0 and 15 (the optimal window length) depending on how many samples in its corresponding segment $\textbf{x}_i$ are artifacts. We visualise the embeddings in 2D using t-SNE, shown in Fig.~\ref{latent_space}.

\begin{table*}[htbp]
    \centering
    \caption{Artifact Detection Performance on the KidsBrainIT dataset. Best performance for each signal is highlighted in bold.}
    \label{tab:combined_performance}
    \begin{tabular}{p{0.9cm}p{1cm}cccccc}
        \toprule
        \textbf{Signal} & \textbf{Metric} & \textbf{VAE-IF} & \textbf{LSTM-ATTn} & \textbf{XGBoost} & \textbf{ARIMA} & \textbf{IF} & \textbf{\makecell{VAE-IF \\ (no attention)}} \\
        \midrule
        \multirow{2}{*}{BPm} & Sens. & \textbf{0.802(0.219)} & 0.767(0.175) & 0.774(0.163) & 0.606(0.238) & 0.294(0.221) & 0.602(0.227) \\
                             & Spec. & 0.921(0.041) & 0.991(0.010) & \textbf{0.995(0.007)} & 0.994(0.006) & 0.973(0.020) & 0.920(0.040) \\
        \midrule
        \multirow{2}{*}{ICPm} & Sens. & \textbf{0.700(0.388)} & 0.695(0.361) & 0.672(0.355) & 0.323(0.316) & 0.274(0.321) & 0.545(0.358) \\
                              & Spec. & 0.926(0.052) & \textbf{0.993(0.007)} & 0.990(0.015) & 0.991(0.006) & 0.959(0.030) & 0.937(0.042) \\
        \midrule
        \multirow{2}{*}{HRT} & Sens. & 0.549(0.425) & \textbf{0.592(0.400)} & 0.465(0.414) & 0.434(0.445) & 0.270(0.366) & 0.395(0.391) \\
                             & Spec. & 0.855(0.227) & 0.916(0.243) & 0.915(0.210) & \textbf{0.922(0.231)} & 0.885(0.255) & 0.863(0.206) \\
        \bottomrule
    \end{tabular}
\end{table*}

\subsection{Ablation study}
We first evaluated the importance of feature extraction using the VAE. For this, we compared the performance of our model with a standard Isolation Forest model trained directly on the patient recordings. Before training, we transformed each sample into a feature vector containing the current sample and 14 previous samples, forming a feature vector of size 15 (i.e., $W=15$ as was used to train the VAE). Next, we analyzed the significance of the self-attention component in the encoder of the VAE. To do so, we removed the self-attention component and trained the model with the same set of hyperparameters estimated by BO. The results presented in Table~\ref{tab:combined_performance}, show that both models achieve low performance compared with the VAE-IF with attention approach.


\section{Discussion}\label{discussion}
In this study, we proposed a fully unsupervised hybrid method leveraging the power of VAE and IF for the detection of artifacts in physiological time series routinely collected at minute-by-minute resolution in ICU settings.  To assess its effectiveness, we conducted a comparative analysis against LSTM-Attention (LSTM-ATTn) and XGBoost, both of which are supervised machine  learnig models, and with a statistical approach using ARIMA. Our results demonstrate that our proposed VAE-IF method achieves sensitivity levels comparable to LSTM-ATTn and XGBoost while maintaining high specificity, with all three models significantly outperforming the standard ARIMA model used for comparison. Notably, our method can operate in circumstances where \textit{no} labels about artifacts are available. Our results clearly demonstrate the ability of VAE-IF to accurately detect artifacts. Furthermore, we visualize the learned latent space representation using t-SNE, revealing that a VAE trained on raw data containing both clean and artifactual samples effectively disentangles clean from noisy instances. These findings highlight the potential of unsupervised methods for artifact detection in physiological time series data, offering a promising solution to the scarcity of labeled data.

In our approach, we employ feature extraction with averaging. Understanding that the latent state of a sample might be influenced by its neighboring samples and their relative positions, we utilized an overlapping sliding-window. When combined with accessing both past and future states via bidirectional LSTM, averaging leads to a more accurate latent representation by taking into account multiple neighborhood contexts.


In comparison to the existing literature \cite{Tsien2001,Mataczynski2022End-to-EndLearning,Subramanian2021UnsupervisedActivity, Chen2021SignalApproaches,Edinburgh2019}, which detects artifacts at segment/window level, our VAE-IF approach detects artifacts on a \textit{sample-by-sample} basis. It is, however, critical to acknowledge that most of these studies focused on waveform data. Here, the primary task involves the identification of deformed morphology in a periodic waveform pulse. In such contexts, it is logical to filter out the entire segment if it is deformed. By contrast, our sample-by-sample strategy is particularly relevant for low-resolution, non-periodic signals. Nonetheless, our approach achieves sensitivity levels similar to XGBoost, a model known for achieving high performance as demonstrated in a previous study \cite{Wu2020}, which also used a sample-by-sample strategy.

Our approach contributes to the existing literature by presenting a compelling case training a VAE on raw data comprising both normal (clean) and abnormal (artifacts) data. In the domain of unsupervised anomaly detection using auto-encoders, which is closely related to artifact detection, the conventional practice entails training the model solely on clean data \cite{Xiong2022AnomalyIForest-AE,Merrill2020ModifiedLearning} and subsequently detecting anomalies during testing by measuring the deviation from the normal data using a distance metric. However, this training strategy becomes counterproductive when labeled data is scarce. In contrast, our approach offers a more practical solution for cleaning ICU data in clinical research. By training the model directly on raw data, we avoid the need for labeled data while still achieving comparable performance. This highlights the potential of our approach as an effective tool for cleaning ICU data, enabling more efficient and reliable clinical research.

In terms of practical utility, multiple factors contribute to the effectiveness of a model for artifact detection. These factors encompass performance, ease of model training, and availability of training examples. Notably, all the models developed in this study exhibit fast training times when utilizing a GPU. Regarding performance, our unsupervised approach demonstrates sensitivity levels comparable to other models, albeit with slightly lower specificity. An additional advantage of our approach is that it does not necessitate any labeled data for training, enhancing its practical applicability.

\subsection{Limitations and future work}

One noteworthy finding of this study is that all models achieved lower sensitivity values when detecting artifacts in the ICPm and HRT signals compared to the BPm signal. A possible explanation for this observation is that a considerable proportion of artifacts in the ICPm and HRT signals share values within the range of normal data, as illustrated in Fig.~\ref{artifacts_dist}. Fig.~\ref{sample_recording} presents an illustrative example of an ICPm recording with an artifact exhibiting the same value range as normal data, illustrating the challenges associated with accurately detecting such artifacts.

The results of our study reveal a potential limitation of our approach, as it demonstrates a slightly lower specificity compared to the supervised models. However, it is important to consider a counter-argument in the context of artifact detection. This lower specificity may still be deemed acceptable, as it could indicate the presence of missed artifacts during the manual annotation process. It highlights the possibility of capturing additional instances that human annotators might have overlooked. Another limitation of our approach is that the model cannot be trained end-to-end, which may hinder the potential for improved performance. Nevertheless, this limitation presents an interesting avenue for future research, specifically in the development of a ``deep'' isolation forest model tailored specifically for artifact detection. Exploring such a direction could potentially enhance the performance and effectiveness of our approach in identifying artifacts with greater accuracy and reliability.

\section{Conclusion}\label{conclusion}
We have successfully developed an unsupervised approach for artifact detection in minute-by-minute physiological recordings, utilizing real-world, routinely collected ICU data. Our approach has demonstrated performance on par with fully supervised methods, highlighting its efficacy and potential for practical implementation as it does not require any data labelling. Furthermore, we have validated the generalizability of our approach by testing it on sample patients from an external dataset, and the model predictions align closely with visual observations. Future studies involving diverse datasets will provide valuable insights and ensure the robustness of our approach to other clinical settings and the possibility to build on VAE-IF to develop an end-to-end method. Overall, our findings present a promising direction for advancing artifact detection methodologies in physiological time series data, with potential applications in clinical research.

\section*{Acknowledgements}
The authors thank Rob Donald for the discussions on artifact detection using the KidsBrainIT-pilot dataset.

This research was funded in part by an EPSRC Doctoral Training Partnership PhD studentship to HH and The Carnegie Trust for the Universities of Scotland (RIG009251) to JE. For the purpose of open access, the author has applied a Creative Commons Attribution (CC BY) licence to any Author Accepted Manuscript version arising from this submission.

\appendix


\bibliographystyle{elsarticle-num} 
\bibliography{references}
\end{document}